\documentclass[11pt,letterpaper]{article}
\usepackage[letterpaper]{geometry}
\usepackage{acl2012}
\usepackage{times}
\usepackage{latexsym}
\usepackage{amsmath}
\usepackage{multirow}
\usepackage{url}
\usepackage{graphicx}
\usepackage{amssymb,amsmath,bm}
\usepackage{textcomp}
\usepackage{subfigure}
\usepackage{caption}

\makeatletter
\newcommand{\@BIBLABEL}{\@emptybiblabel}
\newcommand{\@emptybiblabel}[1]{}
\makeatother
\usepackage[hidelinks]{hyperref}

\graphicspath{{Fig/}}

\def\w{{\bf w}}

\def\z{{\bf z}}

\def\b{{\bf b}}
\def\d{{\bf d}}

\def\k{{\bf k}}

\def\c{{\bf c}}

\def\h{{\bf h}}

\def\wt{\w_t}

\def\U{{\bf U}}
\def\H{{\bf H}}
\def\K{{\bf K}}
\def\Q{{\bf Q}}

\def\bmc{\b_{m_c}}
\def\bmf{\b_{m_f}}
\def\bmi{\b_{m_i}}
\def\bmo{\b_{m_o}}
\def\bl{\b_l}
\def\bc{\b_c}
\def\lm{\d_m}

\title{Recurrent Neural Network Language Model Adaptation Derived Document Vector}

\author{Wei Li \\
 Department of \\ Computer Science and Engineering\\   HKUST, Hong Kong \\
  {\tt wliax@cse.ust.hk} \\\And
  Brian Kan Wing MAK \\
  Department of \\ Computer Science and Engineering \\  HKUST, Hong Kong \\
  {\tt mak@cse.ust.hk} \\
  }

\date{}
\begin{document}
	\maketitle
	\begin{abstract}
		In many natural language processing (NLP) tasks, a document is commonly modeled as a bag of words using the term frequency-inverse document frequency (TF-IDF) vector. One major shortcoming of the frequency-based TF-IDF feature vector is that it ignores word orders that carry syntactic and semantic relationships among the words in a document, and they can be important in some NLP tasks such as genre classification. This paper proposes a novel distributed vector representation of a document: a simple recurrent-neural-network language model (RNN-LM) or a long short-term memory RNN language model (LSTM-LM) is first created from all documents in a task; some of the LM parameters are then adapted by each document, and the adapted parameters are vectorized to represent the document. The new document vectors are labeled as DV-RNN and DV-LSTM respectively. We believe that our new document vectors can capture some high-level sequential information in the documents, which other current document representations fail to capture. The new document vectors were evaluated in the genre classification of documents in three corpora: the Brown Corpus, the BNC Baby Corpus and an artificially created Penn Treebank dataset. Their classification performances are  compared with the performance of TF-IDF vector and the state-of-the-art distributed memory model of paragraph vector (PV-DM).  The results show that DV-LSTM significantly outperforms TF-IDF and PV-DM in most cases, and combinations of the proposed document vectors with TF-IDF or PV-DM may further improve performance. 


	\end{abstract}
	
	
	
	\section{Introduction}
\label{intro}

Document vectorization plays a vital role in many natural language processing (NLP) tasks. One of the most popular document vectors is the \emph{term frequency-inverse document frequency} (TF-IDF) feature vector \cite{robertson1976relevance}. It can offer a robust baseline in many NLP tasks, including genre classification, albeit task specific design may improve the performance. \cite{karlgren1994recognizing,kessler1997automatic,wolters1999exploring,stamatatos2000text,dewdney2001form,lee2002text,freund2006towards,petrenz2009assessing,webber2009genre}.

TF-IDF Document vectorization has a major drawback\cite{cachopo2007improving,le2014distributed}: it ignores word orders and other sequential information in a document. Moreover, TF-IDF and n-gram TF-IDF vectors cannot capture syntactic or semantic relationship/similarity between words, paragraphs, and documents. Besides TF-IDF, another notable document vectorization is the \emph{paragraph vector}\cite{le2014distributed}. Paragraph vectors that learned from a distributed memory model (PV-DM) is a succinct distributed representation of sentences or paragraphs \cite{le2014distributed,dai2015document,ai2016analysis}. PV-DM has been shown to perform significantly better than the bag-of-words model in many NLP tasks. Moreover, Skip-Thought Vectors \cite{NIPS2015_5950} also show superior performances against bag-of-words model.

In this paper, we would like to explore a new document vectorization method that produces a densely distributed representation of text documents while capturing some sequential information in the documents at the same time. Our approach is to adapt/retrain a simple recurrent neural network language model (RNN-LM)
\cite{mikolov2010recurrent} or a long short-term memory RNN language model (LSTM-LM) \cite{sundermeyer2012lstm} with a document, and then vectorize the retrained/adapted model parameters to obtain its document vector (labeled as DV-RNN and DV-LSTM) from RNN-LM and LSTM-LM respectively. The recurrent nature of the RNN-LM or LSTM-LM should capture some high-level and abstract sequential information from its training documents, and our extraction method will not suffer from the limitation imposed by a sliding context window such as in PV-DM. 
In tasks such as text classification or sentiment analysis, different task-specific models have been proposed
\cite{tai2015improved,lai2015recurrent,yang2016hierarchical,zhou2015c,zhang2015character,tang2015document,zhang2016dependency}. 
these models can also take abstract long-term information into consideration. However, these are semi-supervised or supervised methods, which prevent them from training document embeddings from unlabeled data. 
The effectiveness of the proposed document vectors was evaluated on genre classification of documents in three corpora, and their performance was compared with that of TF-IDF vector and PV-DM. Our results show that in most cases, our document vector significantly outperforms n-gram TF-IDF vector and PV-DM.


	\section{Recurrent Neural Network Language Modeling (RNN-LM and LSTM-LM)}
\label{sect:rnnlm}

\renewcommand{\subfigcapskip}{20pt}

\begin{figure}[t]
    \centering
    \includegraphics[width=\linewidth]{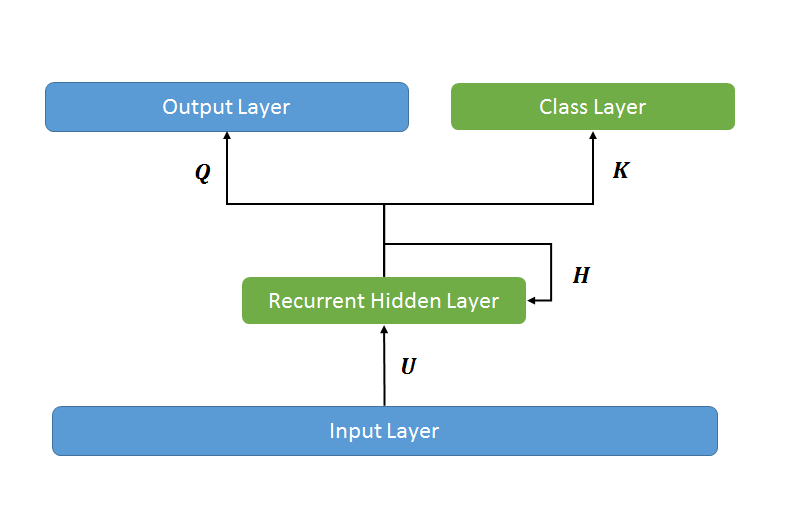}
    \caption{{\it DV-RNN based on a Recurrent neural network language model with output layer factorized by classes.}}
    \label{fig:rnnlm}
\end{figure}

Recurrent neural network language model (RNN-LM) and long short-term memory (LSTM-LM) language model is the state-of-the-art language method \cite{mikolov2010recurrent,mikolov2011extensions,bengio2006neural}. 

Figure~\ref{fig:rnnlm} shows the simple RNN with output layer factorized by classes. In this models, the input is the current word $\wt$ and the output layer is factorized by a (word) class layer. The input word is projected to a distributed representation by the input matrix $\U$. The recurrent matrix $\H$ helps memorize the word history. 


\begin{figure}[t]
    \centering
    \includegraphics[width=\linewidth]{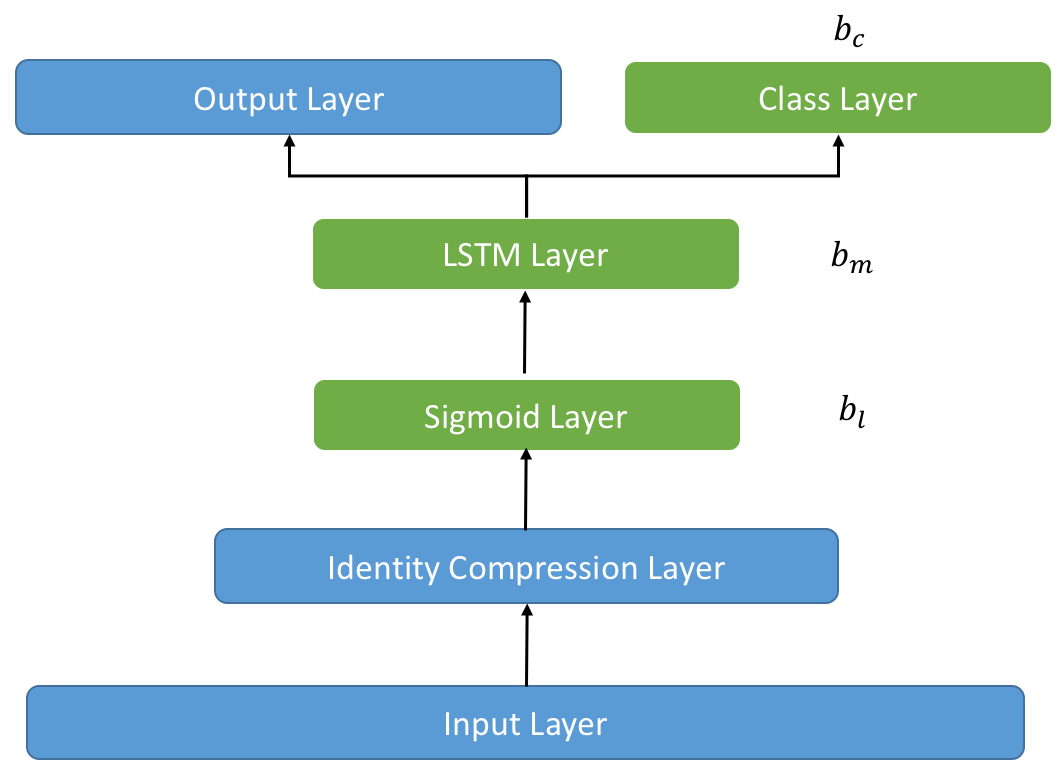}
    \caption{{\it DV-L100-LSTM100 based on a 100-unit-hidden-layer LSTM language model with output layer factorized by classes, with an extra linear compression layer plus an extra sigmoid layer behind the input layer.}}
    \label{fig:DV-LSTM-100}
\end{figure}

Training the above simple RNN may suffer from the problem of exploding or vanishing gradients \cite{hochreiter1997long,sundermeyer2012lstm}. We also propose a network structure for DV-LSTM based on LSTM-LM (labeled as DV-L100-LSTM100, which has a 100-unit hidden layer and a 100-unit sigmoid layer before the hidden layer. (Figure~\ref{fig:DV-LSTM-100}).

	\section{Document Vectorization by RNN-LM / LSTM-LM Adaptation}
\label{sect:dv}

From here on we would refer to our approach as 'adaptation' for simplicity (it can also be considered as retraining). Below is the basic document vectorization procedure: 

\newcounter{step}
\begin{list}{STEP \arabic{step} :}{\usecounter{step}\setlength{\leftmargin 5mm}\setlength{\labelwidth 4mm}\setlength{\labelsep 1mm}\setlength{\itemsep 2mm}}
    
    \item Train a parent LM using all the documents in the training corpus, which serves as the initial model for adaptation.
    \item Adapt the parent LM with each document in the training
    corpus\footnote{Not all model parameters in the parent LM are necessarily adapted. Only the layers shown in green in Figure~\ref{fig:rnnlm}, Figure~\ref{fig:DV-LSTM-100} are adapted in this paper.}
    \item Extract model parameters of interest from the adapted LM, and vectorize them to produce the DV for the adapting document. 
\end{list}

\subsection{Derivation of DV-RNN from RNN-LM adaptation}

As shown in Figure~\ref{fig:rnnlm}, an RNN-LM consists of 4 weight matrices: the input projection matrix $\U$, the recurrent matrix $\H$, the output matrix $\Q$ and the word class matrix $\K$. 
We choose to adapt only $\H$ and $\K$ matrices for the derivation of DV-RNN of a document, while keeping the matrices $\U$ and $\Q$ intact during adaptation. After adaptation, $\H$ and/or $\K$ are vectorized to $\h = \mathrm{vec}(\H)$ and $\k = \mathrm{vec}(\K)$ respectively by enumerate the matrix parameters column by column. The two vectors are concatenated together to form the DV-RNN for the adapting document. 

\subsection{Derivation of DV-LSTM from LSTM-LM adaptation}

The structure of an LSTM cell is more complex than a simple RNN cell, To perform robust LSTM-LM adaptation, we limit ourselves only to adapt the biases in the sigmoid layer $\bl$ (in DV-L100-LSTM100), LSTM layer $\b_m$ and the class layer $\bc$. 
The LSTM biases $\b_m$ is comprised of 4 bias sub-vectors: 
forget-gate biases $\bmf$,
input-gate biases $\bmi$,
output-gate biases $\bmo$,
and cell biases $\bmc$.

%

For DV-L100-LSTM100, all four gates' bias is updated. Plus the $\bl$ and  $\bc$. The final DV-L100-LSTM100 vector, $\d_{m_{100}}$, is the concatenation of the LSTM biases, the sigmoid layer biases and the class layer biases as follows:

\begin{equation}
\d_{m_{100}} = [ n(\b_{a_{100}}'), n(\bc')]' \ .
\label{eqn:dv-lstm100}
\end{equation}

and,

\begin{equation}
\b_{a_{100}} = [n(\bmf'), n(\bmi'), n(\bmo'), n(\bmc'),n(\bl')]' \ .
\label{eqn:ba100.vector}
\end{equation}

\begin{equation}
\b_{m_{100}} = [n(\bmf'), n(\bmi'), n(\bmo'), n(\bmc')]' \ .
\label{eqn:bm100.vector}
\end{equation}

where $\b_{m_{100}}$ is the four LSTM gate biases concatenated in DV-L100-LSTM100. $\b_{a_{100}}$ is the concatenation of $\b_{m_{100}}$ and $\bl$. $n()$ is the normalization operation which would normalize the vector to the unit norm so that the two sub-vectors of very different value range can be concatenated. 



The three different biases in DV-LSTM are supposed to capture various types of information for a document: 
$\bl$ is to capture the abstract and distributed word embeddings from the sigmoid compression layer; 
$\b_m$ is to capture the long-span sequential text information in a document, enabling by the recurrent structure and different gates in LSTM cell;
$\bc$ is to capture the word-class information. 
	\section{Experimental Evaluation: Text Genre Classification}
\label{sect:expt}
  
The DV-RNN, and DV-L100-LSTM100 were evaluated on the genre classification of documents in three corpora: the Brown Corpus \cite{francis1979brown}, the British National Corpus (BNC) Baby dataset \cite{BNCbabyreference}, and an artificially created dataset from Penn Treebank \cite{marcus1995treebank}. 
Performances are reported in weighted F-scores (weighted by the sample number in each category).

\begin{table}[tbph]
  \begin{center}
    \vspace{2mm}

    \caption{\it Some basic information of the corpora in the experiments.}
      \begin{tabular}{|c||c|c|c|}
         \hline
           Corpus & PTB  &  Brown & BNC \\ \hline
           \hline 
           \#Genres&4 &10 &4 \\ \hline
           \#Docs& 239&500 & 182 \\ \hline
	       \#Words& 190K  &1M & 4M  \\ \hline
	       \#Words/Doc& 795 &2K &22K \\ \hline
	       
      \hline
    \end{tabular}
    	    
    \label{tbl:corpora}
  \end{center}	 
\end{table}

\subsection{Corpora used for text genre classification}

Table~\ref{tbl:corpora} summarizes some basic information of each testing corpus. The Penn Treebank dataset(PTB) was artificially extracted from the Penn Treebank Corpus by taking out the documents that have the available genre tags provided by \cite{webber2009genre,PTBgenre}. 
As the number of files in each genre is very unbalanced, we limited the number of documents in each genre to no more than 100 and remove the \emph{errata} genre (as there are very few documents of that genre). We also removed short documents with fewer than 200 words. As a result, the dataset has a total of 239 documents in 4 genres (38 \emph{highlights}, 95 \emph{essays}, 42\emph{letters}, and 64 \emph{news}). For the Brown Corpus, the sub-genres under the \emph{fiction} genre were merged so that the total number of genres was 10 \cite{wu2010fine}.

\subsection{Training of the document vector DV-LSTM}
  
The RWTH Aachen University Neural Network Language Modeling Toolkit (RWTHLM) \cite{sundermeyer2015feedforward,sundermeyer2014rwthlm} was used for training all LSTM-LMs and adapting them to produce the DV-LSTMs. 

The word-classes were determined by the Brown clustering method. 

\subsection{Training of the paragraph vector PV-DM}
\begin{table}[tbph]
	\begin{center}
		\vspace{2mm}
		
		\caption{\it The hyperparameter combinations for fine-tuning PV-DM}
		
		\begin{tabular}{|c | c |}
		
			\hline                        
			Window size  & 5,10,15, 20 \\ 
			\hline            
			Min frequency &0, 5,10,20   \\ 
			\hline
			Negative word samples & 0,10 ,20 \\
			\hline
			Sub-sampling &0, 5E-5  \\ 
			\hline
		
		\end{tabular}
		
		\label{tbl:hyperpara}
	\end{center}	 
\end{table}

The paragraph vectors (PV-DMs)
\cite{le2014distributed} were trained for each document in a corpus using the Gensim toolkit \cite{rehurek_lrec}.  The PV-DMs were trained with an initial learning rate of 0.025, with 20 epochs. It was found that PV-DMs of 500 dimensions provide consistent good performance on all three corpora; they are denoted as $PV_{500}$. One-Tenth close-set data is randomly sampled for fine-tuning, and a grid search for the optimal hyperparameter combination is performed on the each task datasets. The combination of the hyperparameters is the permutation of each parameter value in Table~\ref{tbl:hyperpara}. The respective optimal combinations of the hyperparameters are then chosen for each task to obtain better performance from PV-DM. 



\begin{table}[tbph]
  \begin{center}
    \vspace{2mm}

    \caption{\it the dimension of respective feature vectors.}
      \begin{tabular}{|c|c|c|c|c|c|}
         \hline
          $\c_{500}$ &$\lm$& $\h$ & $\k$ & $\z^{1000}_{5}$&$\z_{5}$ \\ \hline
           \hline 
           500 & 1000& 10000 & 10000 & 1000 & 10000 \\ \hline

      \hline
    \end{tabular}
    	    
    \label{tbl:dimension}
  \end{center}	 
\end{table}

Table~\ref{tbl:dimension} summarizes the dimensions of various feature vectors used for the experiments. $\z^{1000}_5$ and $z_5$ are the TF-IDF feature vector of top 1000 and top 10000 5-gram terms respectively.

\subsection{Experimental Results}
\begin{table}[tbph]
  \begin{center}
    \caption{\it Genre classification performance in terms of F-score. 
    }
    \vspace{2mm}
      \begin{tabular}{| c | c | c | c | c | c | c |} \hline
      	Features & PTB  & Brown & BNC Baby  \\
      	\hline \hline
      	
        1-gram* &- &- & 0.913  \\ \hline
        5-gram*  &- &- &  0.956 \\ \hline
        5-POS* &- &- & 0.947   \\ \hline
        
	5-gram, $\z_5$ & 0.7996&0.6275&0.9623 \\ \hline
	$\c_{500}$ &0.7962&0.6148&0.9407 \\ \hline
	$PV_{500}$ &0.8154&\bf0.6455&0.9820 \\ \hline
	$\b_{m_{100}}$ &0.7559&0.5959&0.9841 \\ \hline
	$\b_{c_{100}}$ &0.8239&0.6326&0.9941 \\ \hline
	$\d_{m_{100}}$  &\bf0.8434&0.6443&\bf1.0000 \\ \hline
$d_{m_{100}}$-$PV_{500}$ &0.8576&0.6613&\bf1.0000\\ \hline
$\d_{m_{100}}$-$\z^{1000}_5$ &\bf0.8607&\bf0.6614&\bf1.0000\\ \hline
    
      \end{tabular}

    \label{tbl:results.fscore}
  \end{center}
\end{table}

The genre classification results using different (combinations of) feature vectors over the three corpora are summarized in Table~\ref{tbl:results.fscore} in terms of weighted F-scores. The bold results represent the best results given by a single feature or a set of combined features. The baseline results labeled with * are quoted from \cite{tang2015automatic,wu2010fine}. 


From the results, we have the following observations:
\begin{itemize}
    \item Our own 5-gram TF-IDF obtained results better than the quoted baseline results from existing works. Thus, we will use our 5-gram TF-IDF results as baselines in the ensuing discussion.

    
    \item Both $PV_{500}$ and our DVs show superior performances comparing to traditional N-gram TF-IDF. 
    
    \item  $PV_{500}$'s performance have shown significant improvements when we conduct the grid search on tuning hyperparameters. 
	DV-LSTM and DV-RNN also have the same hyperparameters as in Table~\ref{tbl:hyperpara} except the window size. Due to the limitation of the current platform, we do not tune these parameters for our DV-LSTMs or DV-RNN. 
	We could also expect a significant room for improvement on DV-LSTM and DV-RNN's performance if such hyperparameters are also fine-tuned like those of PV-DM's.
	
	 \item In most cases, our DV-LSTM's performance is better than PV-DM and PV-DM is better than the 5-gram TF-IDF's performance regarding F-score. The difference is most significant in the PTB dataset and the BNC baby dataset. 
	 To evaluate the statistical significance of the performance difference between each vector, we also adopt the pair sample t-test \cite{dietterich1998approximate}. The confidence threshold is setting at 0.99.
	 All the result highlighted with the bold font is statistically significant, comparing to the performance of 5-gram TF-IDF and PV-DM. And the difference between DV-L100-LSTM100 and PV-DM in the case of the Brown corpus is not statistically significant. Thus, our DV-LSTM show significant improvement over PV-DM and 5-gram vector in most cases, with the exception of  DV-L100-LSTM100 in the Brown corpus's case (where the performance is at least as good as those of PV-DM and the 5-gram TF-IDF).
 
    
    
    \item  The concatenation of $\b_m$ $\bc$ and $\bl$ form the final DV-LSTM vector $\lm$ gives good results than each sub-vectors used along in almost all the cases. Thus, we may conclude that the three bias vectors are complementary to each other for extracting information in genre classification. The gate bias, the class layer bias and the sigmoid layer bias could capture different structural or semantic patterns in the documents.

    \item In all three corpora, combining DV-RNN or DV-LSTM with the 5-gram TF-IDF or $PV_{500}$ gives the best classification performance. This again shows that our proposed document vectors have extracted some information other than term frequency-based information. 
  
    \item The strength of various bias components of our DV-LSTM were investigated. It is interesting to see that the LSTM bias vector $\b_m$ is outperformed by the class bias vector $\bc$. There are two possible reasons: (a) the abstract form of term frequency, which is represented by the class bias parameter, is still the most import feature baseline. (b) DV-LSTM used the Brown clustering method to organize its vocabulary class, which may offer better performance. Thus we also test the performance of the stand along Brown clustering term frequency feature in $c_{500}$ (the same 500 classes as the class bias vector in DV-LSTM), it shows that $c_{500}$ is still much inferior comparing to the 5-gram TF-IDF and $\bc$. Therefore, it could be inferred that it is the recurrent LSTM neural network structure and abstract feature extraction capability in the DV-LSTM makes the greater contribution to the performances.

    \begin{table}[tbph]
  \begin{center}
    \caption{\it Genre classification performance of DV-RNN in terms of F-score.}
    \vspace{2mm}
      \begin{tabular}{| c | c | c | c | c | c | c |} \hline
      	Features & PTB  & Brown & BNC Baby  \\
      	\hline \hline

$\h$-$\k$ &0.8505&0.5836&0.9813\\ \hline
$\h$ &0.7785&0.5156&0.9850\\ \hline
$\k$ &0.7877&0.5743&0.9767\\ \hline
$\h$-$\z_5$ &\bf{0.8622}&0.6343&\bf 0.9936\\ \hline
    
      \end{tabular}

    \label{tbl:results.fscore.rnn}
  \end{center}
\end{table}

    \item  Table~\ref{tbl:results.fscore.rnn} presents the performance of the DV-RNN. In general, DV-RNN's performance is better than PV-DM and 5-gram TF-IDF in the PTB and BNC baby dataset, while worse in the Brown dataset. 

    \item In summary, in terms of weighted F-score, the best results are achieved by the concatenated feature of PV-DM or 5-gram TF-IDF with our DV-L100-LSTM100 vector. Vectors from  DV-L100-LSTM100 achieve the best (or at least equal) single vector performance.
    
\end{itemize}

	\section{Conclusions and Future Works}
\label{sect:conclude}
    
    This paper proposes a novel distributed representation of a document, which we call ``document vector'' (DV). 
    In the current task, DV-LSTM shows good performance when used independently or in conjunction with TF-IDF or PV-DM, and performs better than DV-RNN. Given that DV-LSTM is more compact than DV-RNN and its better performance, DV-LSTM is preferred except that DV-RNNs is faster to train. In the future, we would investigate its effectiveness in other NLP problems such as topic classification and sentiment detection. 
    
    Finally, we believe that the concept of our new DV can be applied to other sequential data (such as gene sequences and stock prices) of variable lengths by treating the sequential data as a document, from which we may create a DV of fixed length to represent them.
    


	\clearpage
	\newpage
	\bibliographystyle{acl2012}
	\bibliography{mybib}


\end{document}